\documentclass[11pt,a4paper]{article}
\usepackage{cite}
\usepackage{amsmath,amssymb,amsfonts}
\usepackage{algorithmic}
\usepackage{graphicx,color}
\usepackage{textcomp}
\usepackage{xcolor}
\usepackage[hidelinks]{hyperref}
\usepackage{algorithm,algorithmic}
\usepackage{array,ragged2e}

\newcolumntype{R}[1]{>{\RaggedLeft\arraybackslash}p{#1}}
\newcolumntype{L}[1]{>{\RaggedRight\arraybackslash}p{#1}}
\newcolumntype{C}[1]{>{\centering\arraybackslash}p{#1}}

\newcommand{\eg}{e.g.~}
\newcommand{\wrt}{w.r.t.~}

\usepackage{multirow}
\usepackage{url}
\usepackage{comment}
\usepackage[belowskip=10pt,aboveskip=10pt]{caption}
\usepackage{subcaption}
\usepackage{booktabs}
\usepackage{tikz}
\usepackage{acronym}
\usepackage{siunitx}

\usepackage[textsize=tiny, colorinlistoftodos]{todonotes}

\usepackage{pifont}
\newcommand{\OK}{\ding{51}~}

\newcommand{\NONE}{$\varnothing$~}

\acrodef{ICP}{iterative closest point}
\acrodef{DOF}{degrees of freedom}
\acrodef{GNSS}{global navigation satellite system}
\acrodef{RTK}{real time kinematics}
\acrodef{UGV}{unmanned ground vehicle}
\acrodef{UAV}{unmanned aerial vehicle}
\newacroindefinite{UAV}{a}{an}
\acrodef{IMU}{inertial measurement unit}
\acrodef{SLAM}{simultaneous localization and mapping}
\acrodef{APE}{absolute pose error}
\acrodef{RPE}{relative pose error}
\acrodef{FOV}{field of view}
\acrodef{EKF}{extended Kalman filter}
\acrodef{APDGICP}{adaptive probability distribution-GICP}
\acrodef{NDT}{normal distributions transform}
\acrodef{DNN}{deep neural network}
\acrodef{CNN}{convolutional neural network}
\acrodef{GICP}{generalized ICP}
\acrodef{FMCW}{frequency modulated continuous wave}
\acrodef{FFT}{fast Fourier transform}
\acrodef{NN}{neural network}
\acrodef{ROS}{Robot Operating System}

\begin{document}

\title{Viking Hill Dataset: A Lidar-Radar-Camera Dataset for Detection and Segmentation in Forest Scenes}

\author{
Vladim\'{i}r Kubelka
\and
Oleksandr Kotlyar
\and
Unal Artan
\and
Martin Magnusson
}

\date{Örebro University, AASS research centre, Robot Navigation and Perception Lab, Örebro, Sweden\\[2ex]28 February 2026} 

\maketitle

\begin{abstract}
Autonomous robots operating under forest canopies need robust perception of trees and surrounding vegetation across varying seasonal conditions.
Existing forestry datasets provide lidar or camera data with per-tree annotations, but none include co-registered 4D imaging radar — a modality of growing interest for its resilience to visual degradation, surface contamination, and vegetation occlusion.
We introduce a multi-sensor forest dataset collected by a mobile robot equipped with a high-resolution FMCW imaging radar, lidar, RGB camera, IMU, and RTK-GNSS.
The site was recorded in two sessions under contrasting vegetation states, and 3D cuboid annotations — including per-tree diameter estimates — provide shared semantic labels across all three perception modalities.
Furthermore, we provide baseline results for semantic segmentation of the radar and lidar point clouds using MinkowskiUNet. 
Radar achieves IoU scores competitive with lidar for dominant classes (ground 91\%, canopy 86\%) while lagging on geometrically fine structures such as tree trunks (56\% vs.\ 74\%). 
A cross-modality analysis further compares lidar and radar trunk segmentation against an RGB detection model, and a diameter-stratified evaluation reveals how trunk segmentation quality varies with tree size.
Beyond segmentation, the co-registered multi-modal data and RTK-GNSS-aided reference positioning support research in mapping, localization, and sensor fusion under canopy. The dataset and annotation tools are publicly available.
\end{abstract}


\section{INTRODUCTION}

Robotic forestry machines require robust perception of trees and surrounding vegetation regardless of their level of autonomy, yet commonly used sensors are susceptible to the dirt, debris, and visual clutter common in forest environments.
\Acp{UAV} are currently a popular platform for forest monitoring as they can quickly cover large areas and even operate under the forest canopy.
They can carry sensors for inventorying and forest health assessment, but a limiting factor is the maximum flight time and the maximum amount of payload~\cite{torresan_forestry_2017}.
\Acp{UGV} complement \Acp{UAV} in applications that require heavy equipment or physical interaction with the environment \cite{oliveira_advances_2021} -- with forms spanning from light, battery-operated robots to automated, diesel-powered forestry machinery.
Recent work has demonstrated autonomous forestry operations ranging from legged robots performing forest inventory~\cite{freissmuth_forest_inventory_2024, chirici_spot_forest_2023} to unmanned forestry machines~\cite{la_hera_autonomous_2024} and autonomous selective harvesting~\cite{rajendran_selective_harvesting_2024}.
Yet, autonomous operation on the ground poses several challenges.
The conditions are harsh; the terrain is often uneven and cluttered with dense vegetation.
The machines and their sensors get covered with dirt, saw dust or vegetation fragments during their operation.
As autonomous operation without human intervention will eventually require a redundant combination of different sensing modalities in the case any of them fail, our work aims at stimulating the research in this direction.
We present a camera-lidar-radar dataset and a semantic segmentation baseline evaluation for \acp{UGV} operating in forest environments, with special focus on an under-represented sensor modality:~high-resolution 4D radars.
The purpose of the dataset is to investigate whether radar contributes useful information in forestry perception.

\begin{figure}
    \centering
    \includegraphics[angle=0, width=0.65\linewidth]{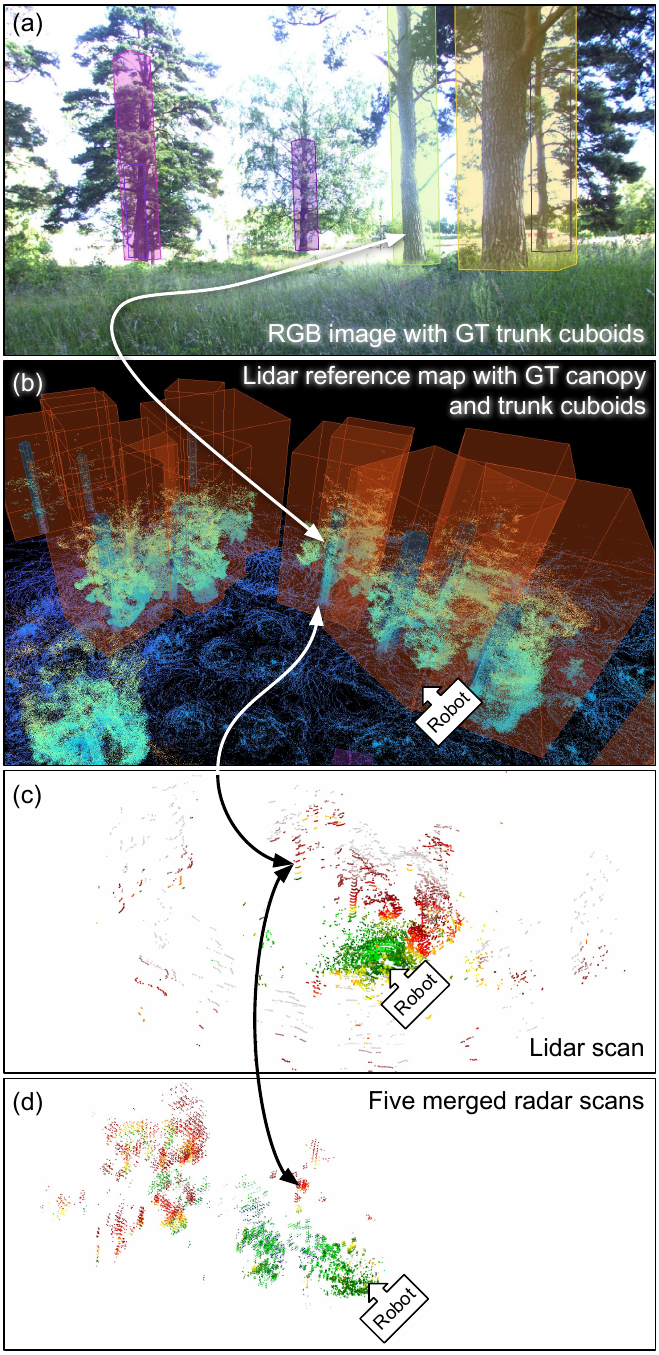}
    \caption{
     Three modalities provided in the dataset: RGB images, lidar, and radar scans. (a) RGB image with projections of one class label -- `tree trunk'; (b) the reference lidar map, with all types of cuboids displayed, showing to the same spot as the camera; (c) 360-degree lidar scan from the same time as the image; (d) five radar scans accumulated in the span of one second at the same time. The point clouds are colored by the elevation coordinate and use the same viewpoint for better interpretability.
    }
    \label{fig:modalities_image}
\end{figure}

The most prominent sensors in this domain have been cameras and lidars.
They are used, for example, for navigation \cite{fortin_uav-assisted_2025, tremblay_topological_2025, baril_kilometer-scale_2021} or in segmentation and detection tasks \cite{oliveira_advances_2021, malladi2025icra}.
In our research, we are interested in modern \ac{FMCW} radars, whose deployment in natural environments is however much less explored.
We are motivated by early experiments performed by Matthies et al.~\cite{matthies_obstacle_2005} with \SI{2.2}{\GHz} radar who were able to detect tree trunks behind two meters of dense undergrowth.
Moreover, Qiao et al. \cite{qiao_radar_2025} has recently demonstrated that path-following in forested area based on a spinning 2D \ac{FMCW} radar is challenging, but very well feasible.
Current research in radar perception for automotive application also shows promising results in low-visibility conditions \cite{sheeny_radiate_2021} and preliminary results from mining applications confirm the ability to penetrate through various atmospheric particulates such as dense smoke, fog, or dust~\cite{skog_human_2024}.
Finally, surface contamination such as mud, dry snow, and dust on the radar enclosure (radome) only partially decreases the signal-to-noise ratio in its measurements, but does not inhibit its function.
For this reason, radars are preferred in applications where sensor contamination is unavoidable \cite{xtonomy2026}.
 
There is a significant number of forestry datasets involving lidars and cameras, \eg recently \emph{DigiForests}~\cite{malladi2025icra}, and an even larger number of lidar-radar-camera datasets for various automotive purposes~\cite{harlow_new_2024}.
While individual studies have presented proof-of-concept experiments detecting obstacles in tall grass and behind foliage~\cite{matthies_obstacle_2005, gusland_imaging_2019}, no existing dataset provides co-registered radar, lidar, and camera data in a forest environment with shared 3D annotations that allow direct comparison of segmentation performance across all three modalities.

\begin{figure}
    \centering
    \includegraphics[angle=0, width=0.90\linewidth]{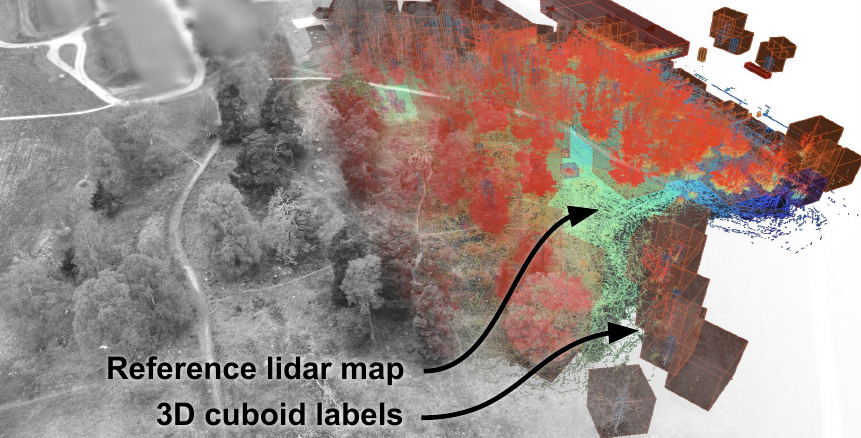}
    \caption{
     The dataset area, mapped by a mobile robot equipped with an imaging radar, lidar and RGB camera. The lidar point cloud map with cuboid labels (overlaid over the aerial photo) serves as a basis for labelling both radar, lidar and camera data into several vegetation classes. 
    }
    \label{fig:title_image}
\end{figure}

One of the possible applications of the dataset (see Fig.~\ref{fig:modalities_image}) presented in this paper is object detection and semantic segmentation for natural environments.
It is motivated by the recent work of Mortimer et al.~\cite{mortimer_goose_2024} and Jiang et al.~\cite{jiang_rellis-3d_2021}.
Beyond segmentation, the synchronized multi-modal data and high-quality ground-truth localization invite further applications such as mapping and terrain analysis.
The dataset captures a forested area in the beginning of May just before a vegetation peak, and in the middle of June with fully developed undergrowth.
The platform used in the dataset was a mobile wheeled robot (Fig.~\ref{fig:robot_sensors}).
A lidar and a pair of \ac{RTK} \ac{GNSS} receivers allowed us to generate a reference point-cloud map and to localize all measurements within its coordinate frame.
This reference map (see Fig.~\ref{fig:title_image}) is accompanied by 3D cuboid labels that allow adding labels into the dataset point clouds and images.
For this purpose, labelling tools have been designed with \ac{ROS} users in mind, allowing easy comparison between lidar, radar and camera as a source of features for detection, segmentation, or SLAM.

In this paper, we provide a radar segmentation baseline, utilizing the state-of-the-art algorithm recently proposed by Malladi et al.~\cite{malladi2025icra} for lidar sensors.
The baseline results indicate that imaging radar is -- at least -- a promising complementary sensor capable of segmenting ground, tree trunks and tree canopies in the point cloud data.
Moreover, the tree trunk segmentation network presented by Grondin~\cite{grondin2022tree} and our camera-lidar-radar cuboid labels allows us to evaluate performance of the lidar and radar segmentation separately on trunks where camera segmentation succeeds and fails.
Additionally, we provide per-tree diameter annotations for the trunk cuboids, enabling evaluation of how segmentation quality varies with tree size — a dimension not captured by aggregate per-class metrics and relevant for downstream tasks where tree diameter determines operational decisions.
In summary, the main contributions of this work are:
\begin{itemize}
    \item First forestry robotics dataset combining lidar, RGB camera, and imaging radar with shared 3D annotations
    \item ROS-based labeling and projection tools
    \item Per-tree DBH annotations
    \item Baseline segmentation results and cross-modal analyses
\end{itemize}

We hope that the presented dataset motivates the robotics community to investigate possibilities of imaging radar sensing in forestry and other natural environments.
The dataset, the accompanying code, and detailed documentation is available at: 
\url{https://github.com/RNP-lab/viking_hill_radar_lidar_camera_dataset}

\section{RELATED WORK}

2D spinning \ac{FMCW} radars have proven to achieve lidar-level accuracy in localization and mapping \cite{harlow_new_2024} and research in this area has been well supported by a series of datasets targetting urban, country or off-road environments \cite{gadd_oord_2024, sheeny_radiate_2021, barnes_oxford_2020, kim_mulran_2020}.
Although this type of radar only provides range and potentially Doppler (radial speed) measurements in a 2D plane, Li et al.~\cite{li_get_2025} have shown that such scans can be segmented into different classes (buildings, vegetation, etc.) and used for improving localization accuracy.
Moreover, Qiao et al.~\cite{qiao_radar_2025} have deployed a spinning \ac{FMCW} radar on a mobile robot and demonstrated radar teach-and-repeat navigation in both structured urban and unstructured forested environments.
In our work, we rather focus on 4D radars, as they provide additional sensing dimension (elevation), useful in environments where mobile robots need to reason about vertical obstacles and motion in all degrees of freedom. 

3D (azimuth, range, and Doppler dimensions) and 4D (extra elevation dimension) beam-forming imaging radars retrieve the direction to the sensed targets not by physically moving the antenna (as in the 2D spinning radars), but instead through comparing phase shifts in signals received by arrays of antennas \cite{harlow_new_2024}.
The majority of radars available for robotic applications have been originally developed for automotive applications, both for classical and autonomous vehicles.
Therefore, it is natural that majority of 4D radar datasets aim at this domain, with specialization in localization and mapping, \eg \cite{kim_hercules_2025, zhang_4dradarslam_2023, choi_msc-rad4r_2023} or object detection, \eg \cite{zhang_dual_2025,  paek_k-radar_2022, palffy_multi-class_2022, zheng_tj4dradset_2022, caesar_nuscenes_2020}, all vital for safe autonomous navigation in mixed traffic.
An exception to this trend is Coloradar~\cite{kramer_coloradar_2022}, a general-purpose 4D imaging radar dataset designed for 6-\ac{DOF} radar-based state estimation and mapping. It addresses the need for radar-based perception in visually degraded environments, such as underground mines, indoor spaces or outdoor environments.
The radars used in these datasets process raw signal samples in a series of steps that involve windowing and \ac{FFT}, to recover the measurement dimensions (range, Doppler, azimuth, elevation).
The result of these operations is a radar tensor (also referred to as a radar cube), from which individual targets are extracted by peak detection.
In some cases, the user of the radar has access to the complete data pipeline, as in the Coloradar dataset \cite{kramer_coloradar_2022}, but often only the radar tensor is available (\eg CARRADA~\cite{ouaknine_carrada_2021}, K-Radar~\cite{paek_k-radar_2022}).
In the case of high-resolution 4D radars, only the individual targets (point clouds) are provided to the user, as in TJ4DRadSet~\cite{zheng_tj4dradset_2022} or, in fact, our case.
This is due to the extensive amounts of raw data high-resolution radars process internally, but also due to the manufacturers' intellectual property protection.

The research toward object detection and semantic segmentation in autonomous driving is extensive, as Yao et al. summarize in their recent review \cite{yao_radar-camera_2024}.
They list 20 different camera-radar datasets and 37 approaches to fusing camera and radar data for the detection and segmentation tasks, all of which are deep-learning based.

In our work, we are interested in field robotics applications, specifically in the forestry domain.
Traditionally, photogrammetry and air-borne lidars have been the sensor modalities of choice for forest biomass monitoring, mainly in the remote sensing and spatial information sciences~\cite{khan_forest_2024, campos_overview_2024, oehmcke_deep_2022}.
Similarly in field robotics, lidars and cameras are the main sensor modalities in currently available forestry datasets.
Tree inventory creation is a major task that can be automated by robotic technology and the works of Tremblay et al.~\cite{tremblay_automatic_2020}, Puliti et al.~\cite{puliti_for-instance_2023}, Cheng et al.~\cite{cheng_treescope_2024}, and Malladi et al.~\cite{malladi2025icra} contribute to that goal with camera and lidar datasets.
They provide ground truth labels for automatic tree segmentation and tree diameter estimation.
The WildScenes dataset from Vidanapathirana et al.~\cite{vidanapathirana_wildscenes_2025}, RELLIS-3D from Jiang et al.~\cite{jiang_rellis-3d_2021} and GOOSE from Mortimer et al.~\cite{mortimer_goose_2024} provide training data for the more general task of semantic segmentation, and they cover more diverse set of natural environments.
None of these datasets includes the radar modality, which we believe is a robust sensor appropriate for machines exposed to the aforementioned harsh conditions.

In our dataset, we provide both lidar and radar as the range-sensing modalities, with the possibility to use the 3D label set for both modalities.
We are motivated by earlier works on developing radars for off-road scenarios: Matthies et al.~\cite{matthies_obstacle_2005} experimented with a single-antenna \SI{2.2}{\GHz} radar in a diffraction tomography setup, demonstrating that a tree trunk can be located behind two meters of vegetation.
Their experiment serves as a proof-of-concept, however their setup would be difficult to replicate on an autonomous machine as it requires a static environment and taking a series of measurements while the radar is being shifted along a defined path.
Moreover, they obtained only a 2D intensity map without considering any classification of the detected objects.
Similarly, Gusland et al.~\cite{gusland_imaging_2019} performed experiments with several objects representing obstacles obscured by tall grass.
Their work however focused mainly on the antenna array design and signal frequency selection and again provided only intensity map with peaks at the object locations.

Besides lidar and radar, our dataset contains calibrated camera data that offer a straightforward way to extract visual features associated with the 3D labels.
We foresee two applications, either developing lidar-radar-camera sensor fusion approaches, or evaluating different combinations of those.
In this paper, we benefit from the work of Grondin et al.~\cite{grondin2022tree} who presented a densely annotated tree-trunk image dataset and trained a deep neural network model for trunk detection, segmentation and key-point extraction. 
We evaluate this model \wrt radar and lidar segmentation performance.
To this end, we also provide results of semantic segmentation of lidar and radar point clouds using MinkowskiUNet-based~\cite{choy20194d} neural network architecture, used  in~\cite{malladi2025icra} for semantic and panoptic segmentation of lidar point clouds of forestry environment from \emph{DigiForests}~\cite{malladi2025icra} dataset.
These results can be considered as a baseline for further evaluation of the dataset by interested researchers and engineers.

\section{DATASET}
\label{sec:dataset}

The dataset captures a forested area near Örebro university, called Enbuskabacken. 
It contains both fully grown and young trees and dense undergrowth.
Two dominant tree species in the area are scots pines and birches.
The terrain is uneven because it used to serve as a burial ground during the Viking Age (550-1050 AD).
A Clearpath Husky robot was teleoperated through the area in May and June 2024, each time driving roughly one hour while recording its onboard sensors: lidar, high-resolution 4D radar, GNSS, RGB camera, IMU and internal odometry. GNSS and lidar measurements were used to generate a reference point cloud map, providing absolute positioning of the robot during the experiment. 

The dataset provides labels (3D cuboids) for several classes (Ground or grass, Tree trunk, Tree canopy, Rock, Bush or small tree, Car, Building or similar, Lamp or sign, Ignore) and \ac{ROS} tools that apply these cuboids to the dataset point clouds (lidar, radar, or accumulated versions of those).
The selected classes are the ones that were possible to distinguish and manually annotate in the point-cloud reference map.
As the motivation for this dataset is primarily detection and segmentation, these classes capture a variety of sizes and geometrical structures visible in all three sensor modalities.
The choice of 3D cuboids as the mean for annotation was motivated by the problem of hand annotation of the radar point clouds. 
While it is practically impossible to distinguish between the defined label classes in the sparse radar point clouds, 3D cuboids created in the reference lidar point cloud map can be easily applied to the radar points.
In Section~\ref{sec:future}, we discuss possible future improvements in the radar-lidar labelling pipeline.
Beside labelling the point clouds, the cuboids can be projected onto the RGB camera images while respecting occlusions, and exported in the COCO dataset~\cite{lin_microsoft_2014} detection format.
All these tools are easily configurable and designed with re-usability in mind.

\subsection{DATA COLLECTION AND REFERENCE LOCALIZATION}

\begin{figure}
    \centering
    \includegraphics[angle=0, width=1.00\linewidth]{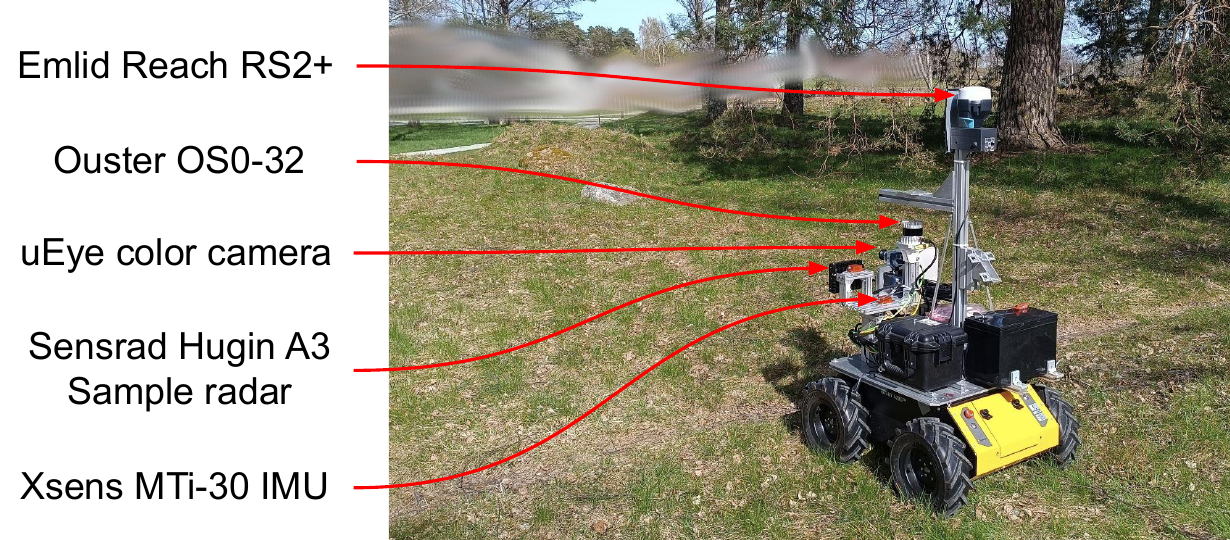}
    \caption{
     Sensor suite on the Clearpath Husky robot. 
    }
    \label{fig:robot_sensors}
\end{figure}

\begin{figure}
    \centering
    \includegraphics[angle=0, width=1.00\linewidth]{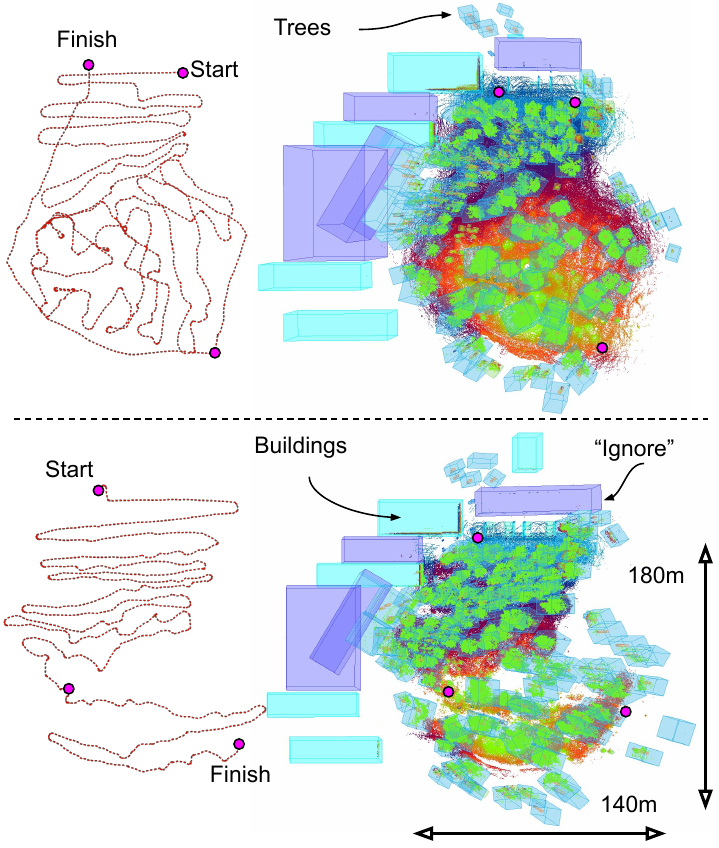}
    \caption{
     Top-down view on the two recording sessions; \emph{short grass} run with its trajectory (top) and its \emph{tall grass} counterpart (bottom). The point cloud is color-coded by height. Note that the \emph{Building} cuboids contain only lidar points on the walls facing the robot during recording, yet we label them by their known size and location. The \emph{Ignore} class is used for areas that contain some lidar or radar points, but we are not able to reliably label them.
    }
    \label{fig:maps_top_down}
\end{figure}

The data were collected during two sessions, in May and June 2024.
In the dataset, we name the sessions \emph{short grass} and \emph{tall grass} to reflect the difference in undergrowth.
The photo in Fig.~\ref{fig:robot_sensors} was taken during the May recording, and besides some younger trees and bushes, the area was quite clear of undergrowth.
On the other hand, the June session captures the vegetation peak, with the undergrowth often reaching above the sensors on the Husky robot.
This provides the opportunity to study the effects of occlusion to the available sensor modalities and specifically evaluate the vegetation penetration capability of the radar sensor. 
The forested area is depicted in Fig.~\ref{fig:maps_top_down}, with the trajectories the robot navigated during the two sessions.
The captured area is approximately \SI{140}{\m} $\times$ \SI{180}{\m}, the maximum elevation difference is approximately \SI{10}{\m} and the terrain is bumpy with overgrown stones as large as the robot itself.

The sensor suite consists of one radar, one lidar, one camera, an \ac{IMU} and a reference RTK-GNSS receiver.
The radar is a Sensrad Hugin A3 radar, with horizontal and vertical \ac{FOV} \SI{80}{\degree} and \SI{30}{\degree} respectively.
Thanks to its configuration of 48 $\times$ 48 transmitting and receiving antennas, the horizontal and vertical resolution is \SI{1.25}{\degree} and \SI{1.7}{\degree}.
The radar was operated in \emph{short range} settings profile, which implies the maximum range is limited to \SI{42}{\meter}, but with a high range resolution of \SI{0.1}{\meter}.
The frame rate was \SI{16}{\hertz} with approximately 3000 to 8000 points per single scan.
The radar has its internal clock set in the beginning of the recording but does not allow NTP or PTP continuous synchronization, some time drift should be therefore expected.
The lidar was an Ouster OS0-32 with the frame rate of \SI{10}{\hertz} and PTP time-stamping.
The coordinate transformation between the lidar and radar was estimated by manually measuring translation between the defined sensor origins.
The rotation was fine-tuned using radar reflectors, observable in both lidar and radar point clouds.
The RGB camera was an IDS Imaging uEye camera with 2056 $\times$ 1542 pixels, calibrated using a chequerboard pattern and the OpenCV camera calibration tool.
Extrinsic calibration followed similar steps as the radar-lidar one, the camera position was manually measured, and the rotation was fine-tuned using point cloud projection onto the camera image and identifying and aligning distinct edges.
Some of the images capture car registration plates and people's faces.
Those have been anonymized using Egoblur~\cite{raina2023egoblur} and Deface\footnote{\url{https://github.com/ORB-HD/deface}}.
The \ac{IMU} was Xsens MTi-30 running at \SI{400}{\hertz} rate, estimating its own attitude using the \emph{VRU General} profile, which does not use magnetometer data to absolutely reference the heading angle, but only to estimate gyro biases and thus limit the heading drift down to \SI{3}{\degree\per\hour} in ideal conditions.
The \ac{IMU} messages are time-stamped by the host computer at arrival -- they can bear some minor time offset caused by the USB transport, but no drift.
All sensor data are available as \ac{ROS}1 and \ac{ROS}2 bag files.

The \ac{RTK} \ac{GNSS} positioning was recorded with a pair of Emlid Reach RS2+ receivers, one serving as a mobile station attached to the robot, the second one served as a reference static station. 
The RTK solution was obtained by post-processing logs of the raw \ac{GNSS} signals using RTKLIB\footnote{\url{https://www.rtklib.com/}}, and the output was added back to the \ac{ROS} bag files, synchronized with the original NMEA messages saved during the recording sessions.
Yet, due to the forest canopy obscuring the sky, even the post-processed RTK solution achieves the highest, sub-centimeter accuracy only at a few spots across the area.
As it is vital to know the robot position with respect to a fixed world frame for efficient 3D labeling of the objects and vegetation, reference lidar maps were created using SLAM  aided by those few high-accuracy GNSS poses.
The SLAM pipeline used the Norlab ICP Mapper\footnote{\url{https://github.com/norlab-ulaval/norlab_icp_mapper_ros}} \cite{Pomerleau12comp} as the front-end for the HDL Graph SLAM\footnote{\url{https://github.com/koide3/hdl_graph_slam}} \cite{koide_portable_2019} optimization back-end. 

The aiding \ac{GNSS} constraints align the map with the Universal Transverse Mercator (UTM) frame. 
Unfortunately, as the number of precise GNSS measurements under the canopy was limited, the absolute accuracy of the reference lidar maps is affected by some drift.
This drift manifests itself as the global map deformation (typically bending).
Each session produced its own map, and when inspecting the alignment of these two maps, we estimate the misalignment at the map limits to $\pm$\SI{30}{\cm}.
Still, the maps are locally accurate and adequate for the intended purpose of the dataset, which is radar and lidar point cloud semantic segmentation training and testing. 
Moreover, the 3D labels -- cuboids -- are provided separately for each session and they align with the sensor measurements.

\subsection{POINT CLOUD LABELLING PROCESS}

The reference lidar maps have been manually annotated by 3D cuboids with the help of online tools from \emph{Segments.ai}\footnote{\url{https://segments.ai}}.
We provide a \ac{ROS} point cloud labelling tool with the dataset which uses these 3D cuboid labels to annotate any point cloud expressed in the map coordinate frame.
The tool adds additional \emph{label\_id} and \emph{cuboid\_id} fields to each point.
The purpose of this design is to provide flexibility for the users; the tools can be re-configured and labels easily recomputed online.
By default, the lidar and radar sensor data can be labelled, but any additional point cloud can be labelled as well.
Moreover, this tool allows setting priority for each label class, such that, \eg tree trunk has priority over tree canopy.
In the current implementation, each point can belong to only one class, and this priority system resolves situations where one point lies inside multiple cuboids.
The tool also offers a way to accumulate several point clouds according to a distance or number threshold and re-publish them as a new, denser point cloud with the possibility to saving to a file as well.
This function has been found useful for the semantic segmentation of radar point clouds, as they typically contain a few thousand points, compared to the nominal 32 thousand points per a lidar point cloud.
Additionally, the dataset contains RGB images from a calibrated camera.
We provide a camera-annotation tool that allows users to choose cuboid classes to be shown in the RGB images and also opt for exporting these in a standard format (COCO dataset \cite{lin_microsoft_2014}) as bounding boxes and contours.

In the dataset, we define nine classes listed in Table~\ref{table:segmentation_performance}, with a special \emph{Ignore} class for areas that contain some lidar or radar points, but where we were not able to reliably label them.
The table also details what the label distribution is in the reference lidar maps.
Since the point density in the reference maps is uniform (unlike the sensor point clouds), similar percentages of class representation can be expected in the lidar and radar point clouds labelled by the \ac{ROS} tool.
As seen from the percentages, the dataset mostly consists of trees.
There are labels for approx. 140 individual trees with distinguishable trunks, and then 50 smaller trees and bushes.


\subsection{PER-TREE DIAMETER ANNOTATIONS}
\label{sec:dbh_annotations}

\begin{figure}
    \centering
    \includegraphics[angle=0, width=1.0\linewidth]{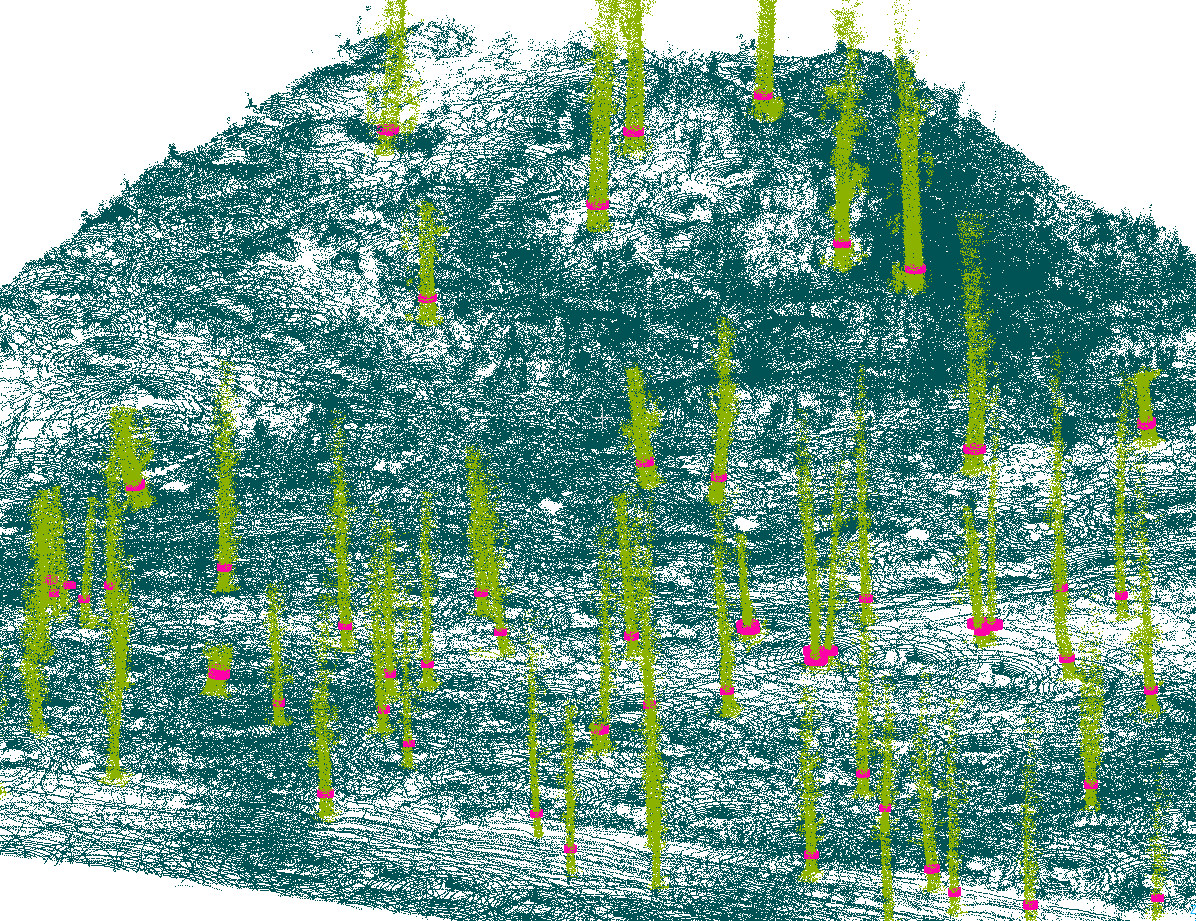}
    \caption{A segment of the reference lidar map from the low vegetation session showing DBH rings (pink) fitted at \SI{1.3}{\m} above the base of each tree trunk (light green). Ground points are shown in dark green.}
    \label{fig:dbh_fit_example}
\end{figure}

To enable evaluation of trunk segmentation as a function of tree size, we provide approximate diameter at breast height (DBH) annotations for the trunk cuboids. Using the cuboid boundaries to isolate individual trunks and the labelled lidar points within them, a circle was fitted to the accumulated lidar points at \SI{1.3}{\m} above the base of each trunk. An automated pipeline performed the initial fitting, followed by manual verification and correction where needed. The DBH values are therefore approximate annotations intended for stratified analysis of segmentation performance rather than ground-truth forestry measurements. The annotations yield 149 trunks in the low vegetation session and 169 in the high vegetation session. For trees with forks where multiple co-dominant stems fall within a single cuboid, the fitted circle encompasses the combined cross-section rather than individual stems. The use of these annotations for stratified segmentation evaluation is presented in Section~\ref{sec:data_analysis}-\ref{sec:trunk_analysis}. Fig.~\ref{fig:dbh_fit_example} shows the fitted DBH rings from a section of the lidar point cloud map.

\section{DATASET APPLICATIONS}
\label{sec:data_analysis}

In this section, we present several complementary analyses that demonstrate how the dataset can be used to evaluate perception methods across sensors, conditions, and object properties. 
We begin with a baseline semantic segmentation evaluation using a MinkowskiUNet model trained separately on lidar and radar point clouds, establishing reference performance and discussing modality-specific effects such as sparsity, ghost reflections, and class imbalance. 
We then perform a cross-modality study conditioned on RGB-based tree trunk detections, analysing how lidar and radar segmentation behave when visual detection succeeds or fails, and how performance evolves with distance. 
Finally, we stratify trunk segmentation performance by tree diameter (DBH), revealing systematic variations across tree sizes and vegetation conditions that are not captured by global IoU scores. 

\subsection{BASELINE SEGMENTATION}
\label{sec:baseline_segmentation}

To deliver a reference point for assessing the performance of artificial neural network-based semantic segmentation on our multimodal dataset, we trained MinkowskiUNet~\cite{choy20194d} for the semantic segmentation of radar and lidar point clouds. Minkowski Convolutional Neural Networks (CNN) are designed to efficiently process spatially sparse data, such as lidar or radar point clouds, by performing convolutions only at occupied, non-zero locations. 
It was shown in~\cite{malladi2025icra} that MinkowskiUNet outperforms other approaches for semantic segmentation of lidar point clouds in forests. We have adopted their neural network architecture implementation with semantic segmentation head, from publicly available code\footnote{\url{https://github.com/PRBonn/digiforests/}}. 
As the input to the neural network, we used the spatial coordinates of the point clouds as the \emph{coordinates}.
Lidar intensity and radar distance-normalized power were used as \emph{features} for lidar and radar point clouds, respectively.
The distance-normalized power represents a rough equivalent to radar cross section (RCS), where we only compensate the distance-related loss:
\begin{equation}
P_{\text{comp}} 
= P_{\text{received}} + 20 \cdot \log_{10}(4\pi) + 40 \cdot \log_{10}(R)
.
\end{equation}
Here, $P_{\text{comp}}$[dB] is the indicated power value in the radar scan and $R$[m] is the range.

\begin{figure}
    \centering
    \includegraphics[angle=0, width=0.6\linewidth]{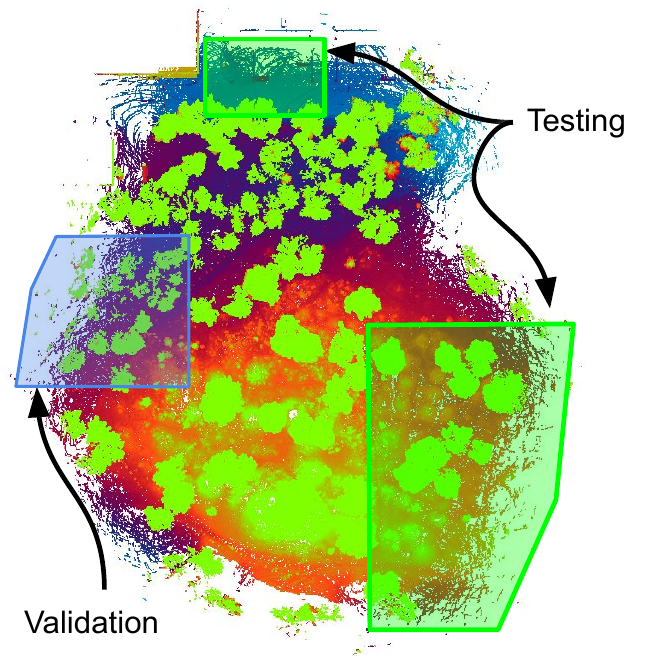}
    \caption{
     The training-validation-testing data split for the semantic segmentation baseline. The same regions were used in both recording sorties.
     The point clouds captured from the marked areas were used for the corresponding data split sets, the training set is implicitly all the remaining point clouds. The point cloud is color-coded by height.}
    \label{fig:data_split}
\end{figure}

In this baseline segmentation application, the dataset was split with the following proportions: 77\% for training, 10\% for validation and 13\% for testing. The dataset split is visualised in Fig.~\ref{fig:data_split}, showing that the subset for the model testing was taken from the areas that have not been used for training.
The training was performed using all nine classes listed in Table~\ref{table:segmentation_performance}.  
We would like to mention that the presence of underrepresented classes -- in our case  \emph{Rock}, \emph{Car}, \emph{Building}, \emph{Lamp or sign}, and \emph{Ignore} -- can lead to class imbalance problems, such as biased learning for the majority and poor generalization for the minority classes. 

The segmentation results for all classes on the test set for lidar and radar point clouds are presented in Table~\ref{table:segmentation_performance}. 
The table provides results for two realizations of the radar training and testing data: \emph{Radar 1-scan}, and \emph{Radar 5-scan}.
They stand for using point clouds consisting of a single radar scan, or five scans accumulated over approximately one meter of robot travel, respectively.
The lower point count in single radar scans is compensated by the 5-scan aggregation spanning one second, resulting in a point count comparable to the lidar.
That is, our lidar produces scans consisting of 33 thousand points, a single radar scan contains up to approximately 8 thousand points.

The best scores for semantic segmentation are as expected achieved on lidar point clouds across all classes in the dataset, confirming the findings in DigiForests~\cite{malladi2025icra}.
At the same time, for the most represented classes in the dataset, \emph{Ground or grass}, \emph{Tree trunk}, and \emph{Tree canopy}, radar point clouds achieve IoU scores within a reasonable range of lidar performance. 
The main source of error are occasional patches of ground labelled as tree trunks and ghost reflections caused by a tarmac road in the north part of the map, where mirrored tree trunks and canopies appear below ground and the model labels them based on their shape as one of those classes.

\begin{figure}
    \centering
    \includegraphics[angle=0, width=1.0\linewidth]{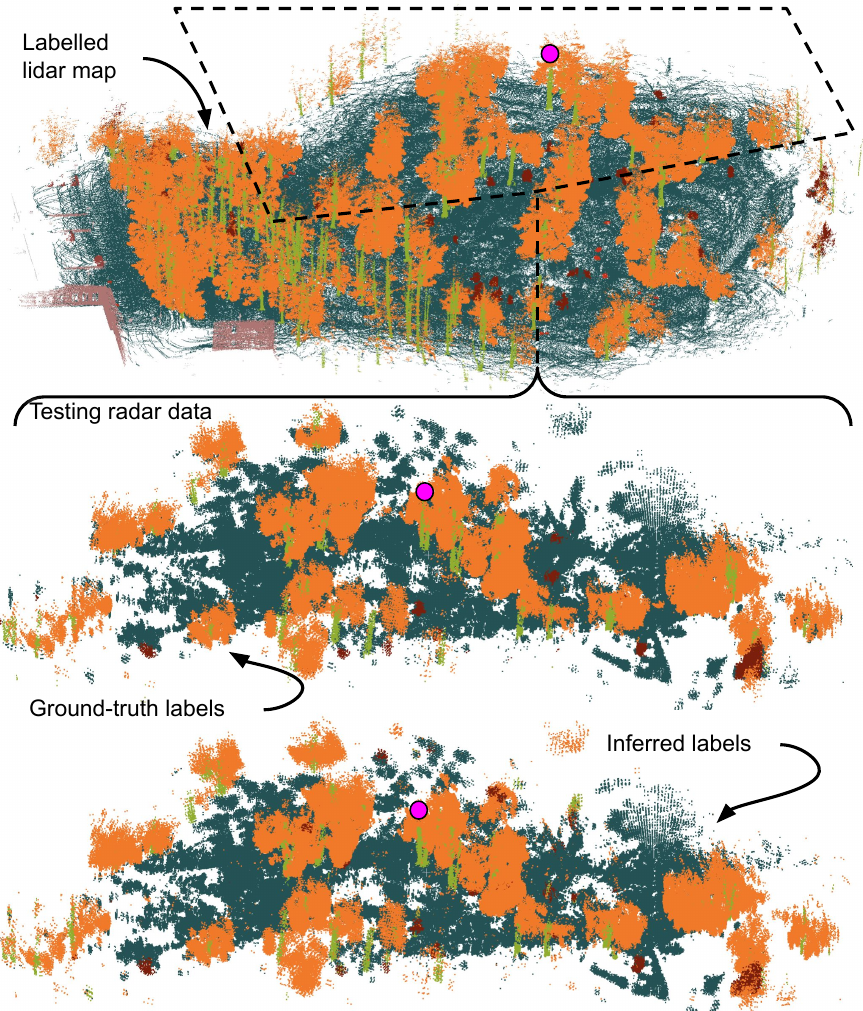}
    \caption{
     Visualization of the inference results on the radar testing set. For reference, the lidar map is shown at the top, with an approximate region marked from which the testing radar data are drawn (shown below). The point clouds are colored according to different classes (orange -- tree canopy, dark green -- ground, light green -- tree trunks, brown -- bushes). The purple dot denotes the same tree, and all three point clouds are seen from the same view point for easier orientation.}
    \label{fig:radar_inferrence}
\end{figure}

Fig.~\ref{fig:radar_inferrence} shows qualitative results on a test part of the radar dataset. 
The results of the MinkowskiUNet model prediction (low panel) is visually indistinguishable from the ground truth (middle panel) for the classes \emph{Ground or grass} and \emph{Tree trunk}, some minor difference can be seen for the \emph{Tree canopy} class, and some distinguishable differences can be observed only for \emph{Bush or small tree} class.
Segmentation of bushes gives the lowest performance among all major classes in both the lidar and radar datasets.
This is presumably due to the fact that patterns learned by the model for the trees can also be present in bushes, and thus the model misclassifies bushes as trees.
Additionally, we would like to emphasize that for the class \emph{Lamp or sign}, the lidar segmentation performance achieves around 60\% which shows that the trained model learned meaningful representations for this class, contrary to the radar counterpart, which mostly labels it as the \emph{Tree trunk} class.

Our work training the semantic segmentation model on this dataset raised several interesting questions.
The radar scans typically tend to make narrow structures such as the tree trunks seem thicker.
This effect is related to the radar resolution and settings (range resolution is \SI{10}{\cm}), specifically how the radar identifies targets in the radar tensor.
In our experiments, we opted for adjusting the corresponding cuboid sizes such that the ground-truth labels would include the additional points, which however lie outside the physical tree trunk.
Moreover, the radar scans often contain noise points, either caused by reflections on flat surfaces, or by multipath interference.
It is left for discussion if these need a special label class -- in our experiments, we left them marked as \emph{Ground or grass} (the default background class).
Finally, although our dataset is mostly static, the Doppler measurements provided by the radar could bear some information relevant to the semantic segmentation.
However, the Minkowski CNN training can involve random augmentations, and the Doppler values would require specific handling to remain consistent with the transformed velocity vector.
As the Doppler velocity is measured in the sensor frame, geometric augmentations change the relationship between the velocity and the coordinate frame.
Providing correct transformation was outside the scope of the baseline and we leave this problem for future work.

\begin{table}
     \centering
     \begin{tabular}{l|R{0.15\linewidth}R{0.09\linewidth}R{0.13\linewidth}R{0.13\linewidth}}
     \toprule
     & Label distribution [\%] & Lidar IoU[\%] & Radar 1-scan IoU[\%]  & Radar 5-scan IoU[\%]\\
     \midrule

     Ground or grass    & 38.89 & 98.4 & 89.3 & 90.5 \\
     Tree trunk         & 8.73 & 74.3 & 38.8 & 55.8 \\
     Tree canopy        & 48.04 & 91.0 & 80.5 & 85.5 \\
     Rock               & 0.12 & 61.3 & 0.0 & 0.0 \\
     Bush or small tree & 3.12 & 52.1 & 29.0 & 38.9 \\
     Car                & 0.16 & 42.1 & 0.0 & 18.5 \\ 
     Building           & 0.84 & 65.7 & 27.0 & 29.7 \\    
     Lamp or sign       & 0.07 & 41.2 & 0.0 & 0.11 \\
     Ignore             & 0.02 & --   & --  & --   \\

     \bottomrule
     \end{tabular}
     \caption{Label distribution in the dataset and segmentation performance of MinkowskiUNet on the test set for lidar and two realizations of radar point clouds. Results are reported in terms of Intersection over Union (IoU) scores and given in percentages. }
     \label{table:segmentation_performance}
\end{table}

\subsection{SEGMENTATION CONDITIONED BY RGB DETECTION}

As the dataset allows projecting the 3D cuboids into the RGB images and Grondin et al.~\cite{grondin2022tree} provide a pre-trained, state-of-the-art tree trunk detection and segmentation model, we can perform a cross-modality performance analysis. 
More specifically, we are interested in the behaviour of the lidar and radar in the cases when the visual model succeeds in detecting a tree trunk, and also when it fails.
Our assumption is that the missed visual detections are caused by unfavourable light conditions, such as under- or over-exposed parts of the camera image, but also by other vegetation partially or completely occluding the view.
The missed visual detections are especially interesting, as they allow us to quantitatively test assumptions about radar and lidar penetration through the undergrowth.

\begin{figure}
    \centering
    \includegraphics[angle=0, width=1.0\linewidth]{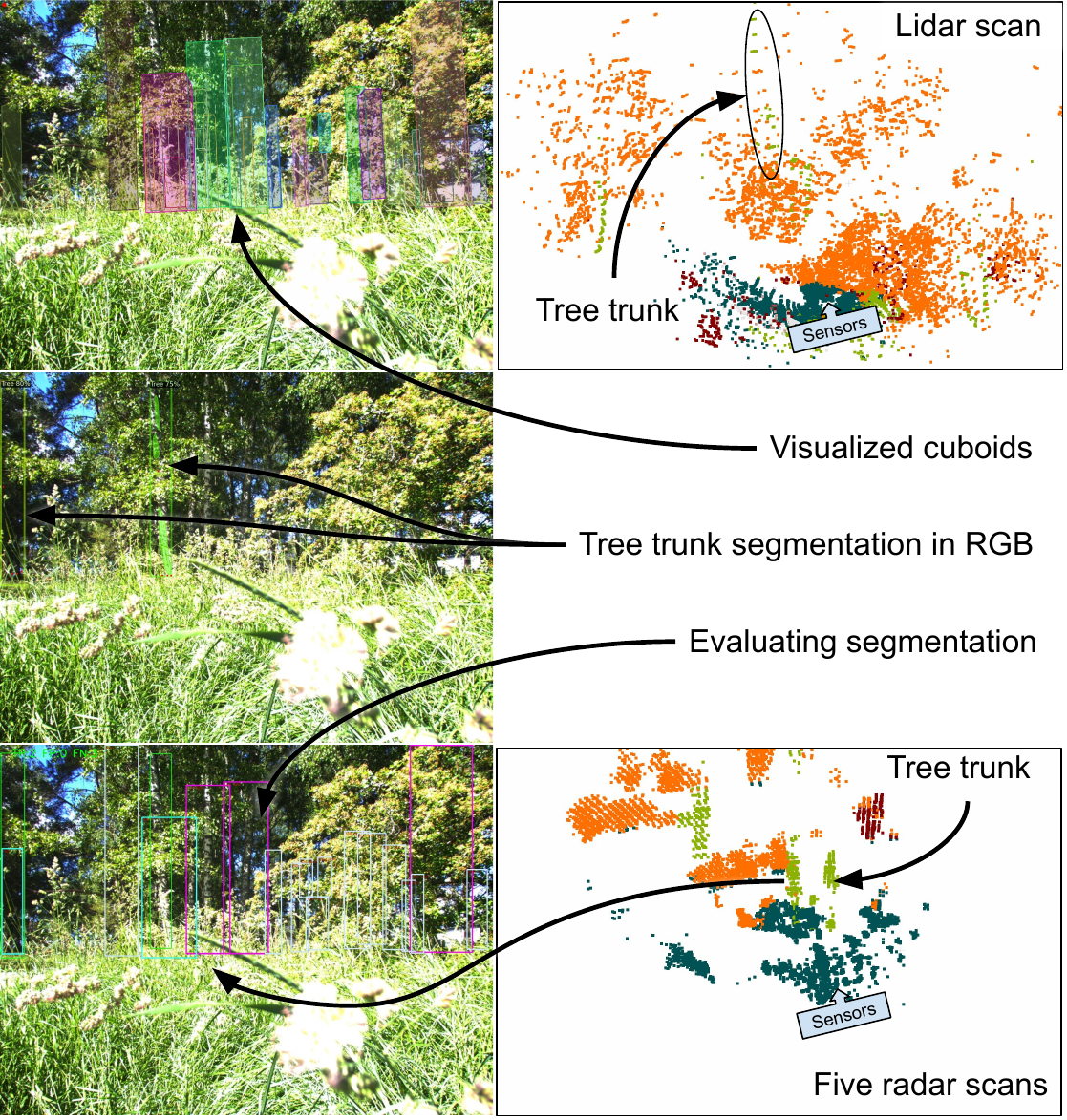}
    \caption{
     Visualization of steps in the tree-trunk detection cross-modality evaluation. The cuboid labels are projected into the RGB image (top left) with ranges and occlusions resolved, then matched with RGB segmentation (middle left) and point-cloud segmentation (top and bottom right).}
    \label{fig:rgb_lidar_radar_eval}
\end{figure}

We set up the analysis as follows: The visual model had been trained on the dataset documented in \cite{grondin2022tree}, the specific property of that dataset is that only trees up to \SI{10}{\m} had been labelled.
Their motivation is that this detection and segmentation model should serve for autonomous harvesting operations, where ten meters is the usual reach radius of the harvester crane.
We therefore need to limit our initial analysis only to a distance bin \SI{0}{}-\SI{10}{\m}.
The evaluation pipeline is described in Fig.~\ref{fig:rgb_lidar_radar_eval}.
The 3D cuboids are projected into the RGB image, and their occlusions are resolved as follows: cuboids of class \emph{Car} and \emph{Building} mask the class \emph{Tree trunk} and the tree trunk cuboids mask each other based on distance from the camera.
Using the painter's algorithm, the final tree trunk cuboid masks are painted onto the RGB image.
To avoid cluttering the image with small fragments, we discard masks whose un-occluded portion is less than half of their original size.
The remaining masks are converted to 2D boxes for which we store the original cuboid depth in the camera view, class and cuboid id.
Subsequently, inference using the visual model is executed on the original RGB images and the detections are matched to our 2D boxes.
If a tree trunk detections horizontally overlaps more than 50\% with a cuboid box, the sum of true positive detections is increased by one.
We decide not to penalize true positive (TP) detections further than \SI{10}{\m}, these are also counted as TP.
Cuboid boxes within \SI{10}{\m} without a detection match are considered false negative detections (FN).
Tree trunk detections without a matched cuboid box are considered false positives (FP).
The results of this process are shown in the first five columns of Table~\ref{table:rgb_lidar_radar}.
Precision and recall confirm the results reported by \cite{grondin2022tree} and we also observe the expected effect of high vegetation - the recall in that case decreases by approximately 8\%.
Visual inspection of the FN cases has revealed that -- especially in the low vegetation case -- a significant portion these misclassifications is caused by tree trunks that are partially or even completely outside the image frame, while portions of the cuboid are still visible.
We discuss this problem further in Section~\ref{sec:future}.
In the high vegetation conditions, vegetation occlusions and also higher contrast due to sunny weather is the major contributor to the FN value for visual classification.
In both sessions, the main source of FP detections are lamp poles and similar structures and trunks of young trees that we label separately (their trunks are not visible in the reference lidar map).

We proceed with analysing the performance of the lidar and radar segmentation, based on our trained Minkowski convolutional network.
Inspecting only the cuboids within the view of the RGB camera, we split these into two groups; those who correspond to the TP RGB detections, and those which belong to the missed FN detections.
Statistics capturing the point segmentation inside these cuboids are shown in the rest of the Table~\ref{table:rgb_lidar_radar} columns.
The symbol \OK stands for cuboids, where more than 70\% of the points were labelled by the network as the tree trunk class.
This threshold corresponds to the 70\% confidence threshold in the visual model.
\NONE denotes cases where the sensor provided no points that fall within the cuboid.
Recall (Rec.) is the median value of the correctly labelled and all observed points in a cuboid, considering all cuboids from the group.
It is important to keep in mind that the sizes of the TP and FN groups are highly disproportionate due to the high RGB model recall.

In these numbers, we observe several trends.
In the low vegetation season, we see a high number of empty measurements, regardless of sensor type or the TP or FN RGB split.
This is generally explainable by the fact that the missing undergrowth and foliage results in less detected points.
Moreover in lidar, we observe consistent numbers of detected trunks and median recall both in the RGB TP and FN splits, and in both seasons.
It is not surprising, as \SI{10}{\m} is rather a short distance for a modern high-resolution lidar.
Due to the distance limit posed by the visual model, we do not observe too many instances in our data where a bush or smaller tree would occlude tree trunks behind it.
In the case of the radar, we observe a substantial decrease of sufficiently labelled tree trunks in the RGB FN split.
In the low vegetation session, we attribute this phenomenon to the fact that a significant number of FN tree trunks are located partially or fully outside the camera frame.
Due to the short distance, this puts the trunks on the edge of the azimuth range of the radar, where less points are detected (almost one fifth of the frames are completely empty).
We see an improvement during the high vegetation session with higher number of detected and correctly labelled points, yet the radar modality still only reaches  half of the lidar performance in these conditions.

\begin{figure}
    \centering
    \includegraphics[angle=0, width=1.0\linewidth]{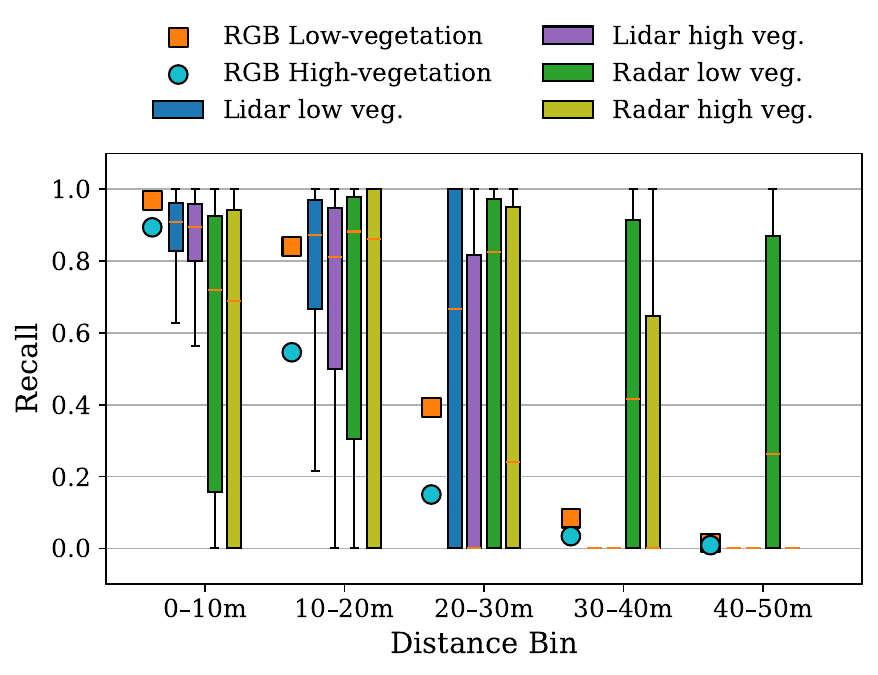}
    \caption{
        RGB image, lidar and radar detection recall evaluated in five distance bins. The box plots denote the minimum, the maximum, the sample median, and the first and third quartiles. In the final two bins between \SI{30}{\m} and \SI{50}{\m}, the lidar used in our experiments was out of range due to generally worse reflectivity of surfaces in the forest. Also note that the \emph{PercepTreeV1} had been trained on trunks annotated up to \SI{10}{\m}.}
    \label{fig:lidar_radar_rec_vs_dist}
\end{figure}

Until this point, only the \SI{0}{}-\SI{10}{\m} distance region has been analyzed.
By increasing the considered range up to the sensor limits and including all cuboids within the camera view, we obtain results shown in Fig.~\ref{fig:lidar_radar_rec_vs_dist}.
Since the visual model has been trained on trunks annotated up to \SI{10}{\m}, we see a quick decline in its performance behind this range.
In the case of the lidar point cloud segmentation, we observe a slightly slower decline, but this sensor also reaches its limits in the \SI{20}{}-\SI{30}{\m} distance bin.
Within its operational range, the lidar segmentation shows smaller spread of recall values than the radar counterpart.
On the other hand, the 4D radar segmentation is able to detect tree trunk points up to \SI{50}{\m} in the low vegetation season.
The spread of the recall values is much higher throughout the whole range.
We explain this fact by the relatively low spatial resolution of the radar sensor, compared to lidars.
As shown in Fig.~\ref{fig:rgb_lidar_radar_eval}, bottom right, a tree trunk is represented by only a few points.
Missing any of these has substantial effect on the segmentation performance compared to the lidar alternative, where the model receives approximately ten time more points.

\begin{table}[t]
\centering
\small
\setlength{\tabcolsep}{2pt}  

\resizebox{\textwidth}{!}{%
\begin{tabular}{
l
*{1}{C{0.9cm}}
*{4}{C{0.7cm}}
*{3}{C{0.8cm}}
*{3}{C{0.8cm}}
*{3}{C{0.8cm}}
*{3}{C{0.8cm}}
}
\toprule

& \multicolumn{5}{c}{RGB: PercepTreeV1}
& \multicolumn{6}{c}{Lidar: MinkowskiUNet}
& \multicolumn{6}{c}{Radar: MinkowskiUNet} \\

& \multicolumn{5}{l}{}
& \multicolumn{3}{c}{RGB TP}
& \multicolumn{3}{c}{RGB FN}
& \multicolumn{3}{c}{RGB TP}
& \multicolumn{3}{c}{RGB FN} \\

\cmidrule(lr){2-6}
\cmidrule(lr){7-9}
\cmidrule(lr){10-12}
\cmidrule(lr){13-15}
\cmidrule(lr){16-18}

& TP & FP & FN & Prec. & Rec.
& \OK & \NONE & Rec.
& \OK & \NONE & Rec.
& \OK & \NONE & Rec.
& \OK & \NONE & Rec. \\

\midrule

Low-vegetation  & 19300 & 949 & 545 & 0.95 & 0.97 & 67\% & 14\% & 0.86 & 74\% & 16\% & 0.87 & 63\% & 14\% & 0.84 & 32\% & 19\% & 0.03 \\
High-vegetation & 12451 & 540 & 1379 & 0.96 & 0.9 & 74\% &  4\%  &  0.86  &  78\%  &  3\%  &  0.88 & 60\% & 7\% & 0.84 & 36\% & 8\% & 0.29 \\

\bottomrule
\end{tabular}
}
    \caption{Performance comparison across sensing modalities for the \emph{Tree trunk} class. RGB camera detection performance is evaluated in \SI{10}{\meter} range according to the original network training process~\cite{grondin2022tree}. Lidar and radar segmentation is evaluated in two splits, following the true-positive and false-negative RGB detections. \OK stands for the number of tree trunks from the given split, where at least 70\% of points were correctly labelled. \NONE denotes the number of tree trunks in the split with zero points detected by the sensor. \emph{Recall} for radar and lidar is defined as the median ratio of correctly labelled points and all points detected by the sensor.}
    \label{table:rgb_lidar_radar}
\end{table}

\subsection{TREE TRUNK SEGMENTATION BY DIAMETER}
\label{sec:trunk_analysis}

The baseline segmentation evaluation (Section~\ref{sec:data_analysis}-\ref{sec:baseline_segmentation}) reports a single IoU score per class, aggregated over all instances within a designated test area. For the \emph{Tree trunk} class, which is central to forestry applications such as inventory creation and selective thinning, this aggregate view can mask potentially significant variation across individual trees. Using the per-tree DBH annotations described in Section~\ref{sec:dataset}-\ref{sec:dbh_annotations}, we stratify the MinkowskiUNet lidar trunk segmentation by tree diameter, evaluating across all lidar scans in each session rather than the test area alone.

\begin{figure}
\centering
\includegraphics[width=1.0\linewidth]{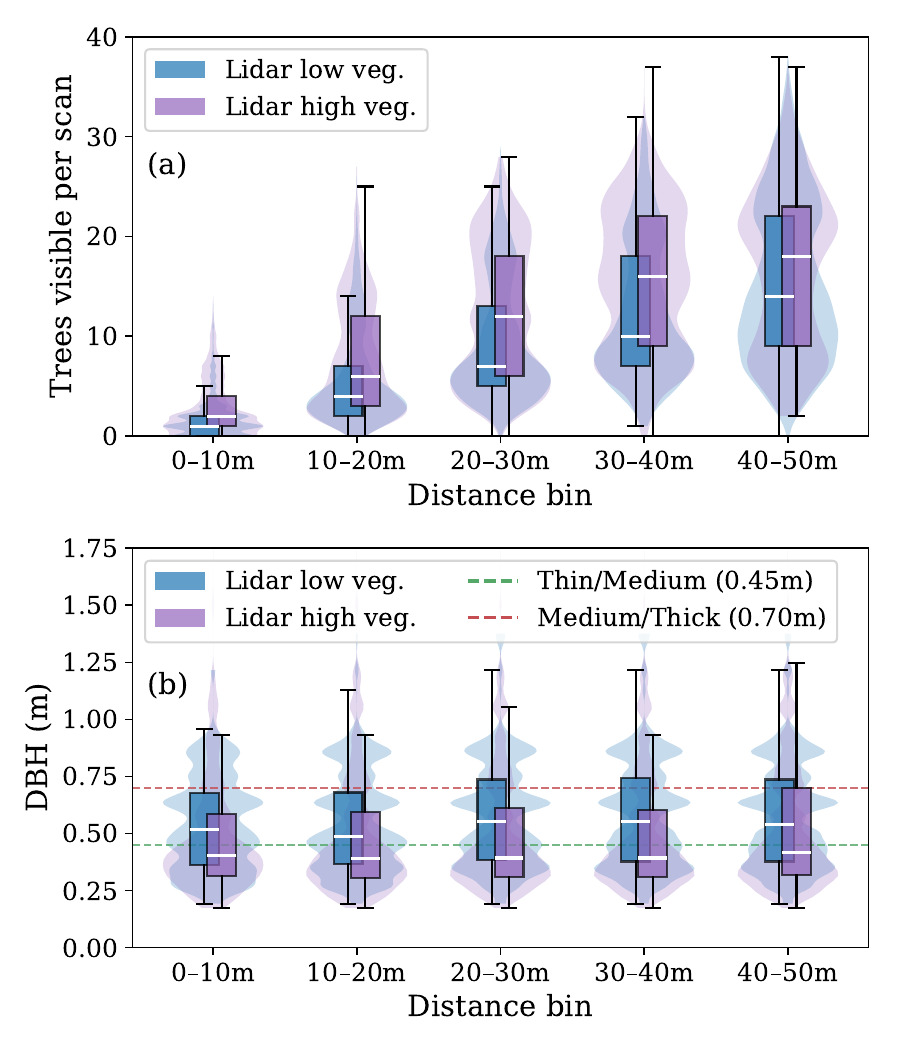}
\caption{Distribution of annotated trees visible per lidar scan as a function of distance from the scanner. The box plots denote the sample median and the first and third quartiles. (a)~Number of trees visible per scan; the count increases with distance as the scanner's field of view encompasses more of the forest. (b)~DBH distribution at each distance bin; the diameter mix remains relatively uniform across bins for both sessions.}
\label{fig:tree_population}
\end{figure}

We group trees into three diameter classes: thin (DBH $<$ \SI{0.45}{\m}), medium (\SI{0.45}{\m} $\leq$ DBH $<$ \SI{0.70}{\m}), and thick (DBH $\geq$ \SI{0.70}{\m}). Fig.~\ref{fig:tree_population}(a) shows the number of visible trees per scan as a function of distance from the scanner; the count grows with distance as the scanner's field of view encompasses more of the forest. Fig.~\ref{fig:tree_population}(b) shows the DBH distribution at each distance bin, which remains relatively uniform — confirming that the three DBH classes are not concentrated at particular distances. For each tree with a DBH annotation, we compute the per-ray intersection-over-union (IoU) across all scans in which it is visible.

Table~\ref{table:trunk_by_dbh} presents the results. In the \emph{low vegetation} session, mean per-tree IoU is similar for thin and medium trees (65.1\% and 61.7\% respectively) but drops for thick trees (55.0\%). The \emph{high vegetation} session shows the same trend more pronounced: 57.8\% (thin), 47.4\% (medium), 44.1\% (thick), with all values lower than their \emph{low vegetation} counterparts — which could be attributed to the taller vegetation in this session.

The lower IoU for thick trees is notable, as larger trunks present a bigger target. Mean observation distance is similar across diameter classes (approximately \SI{11}{\m} in \emph{low vegetation} and \SI{9}{\m} in \emph{high vegetation}), suggesting distance is unlikely the cause. Possible factors include canopy and branch occlusion, cuboid annotations encompassing non-trunk structure such as forks, and underrepresentation of thick trees in training data.

These results demonstrate that aggregate per-class trunk IoU — 74.3\% for lidar on the test area (Table~\ref{table:segmentation_performance}) — does not reflect uniform performance across all trees. We present this diameter-stratified analysis for the lidar modality, where the cuboid boundaries most closely correspond to the physical trunk extent. The diameter annotations enable this stratified evaluation for any segmentation method or sensor modality.
\begin{table}
\centering
\small
\begin{tabular}{l c c c c}
\toprule
& \multicolumn{2}{c}{Low veg.} & \multicolumn{2}{c}{High veg.} \\
\cmidrule(lr){2-3} \cmidrule(lr){4-5}
DBH class & Trees & IoU [\%] & Trees & IoU [\%] \\
\midrule
Thin ($<$0.45\,m)        & 53 & 65.1 & 90 & 57.8 \\
Medium (0.45--0.70\,m)   & 52 & 61.7 & 37 & 47.4 \\
Thick ($>$0.70\,m)       & 44 & 55.0 & 42 & 44.1 \\
\midrule
All                       & 149 &      & 169 &      \\
\bottomrule
\end{tabular}
\caption{Per-tree trunk segmentation IoU (MinkowskiUNet, lidar) stratified by diameter at breast height. Values are means over all scans in which each tree is visible.}
\label{table:trunk_by_dbh}
\end{table}

\section{DISCUSSION ON FUTURE DATASETS}
\label{sec:future}

In future work, we plan to expand the dataset to cover a wider range of environments and seasonal conditions.
To confirm our assumption about sensor robustness against sensor surface contamination, experiments with saw dust and dirt splash will be performed given the robot platform and its sensor suite are fully dust- and water-proofed.
The sensor suite is being extended by two additional Hugin radars to provide \SI{270}{\degree} horizontal field of view.
Moreover, worker safety near autonomous machines is a concern for all industries.
Future datasets will explicitly include people in the captured objects.
For detecting and segmenting those, exploiting radar Doppler information is a promising direction, which however requires proper handling in the training process due to the platform ego-motion.

\begin{figure}
\centering
\includegraphics[width=0.9\linewidth]{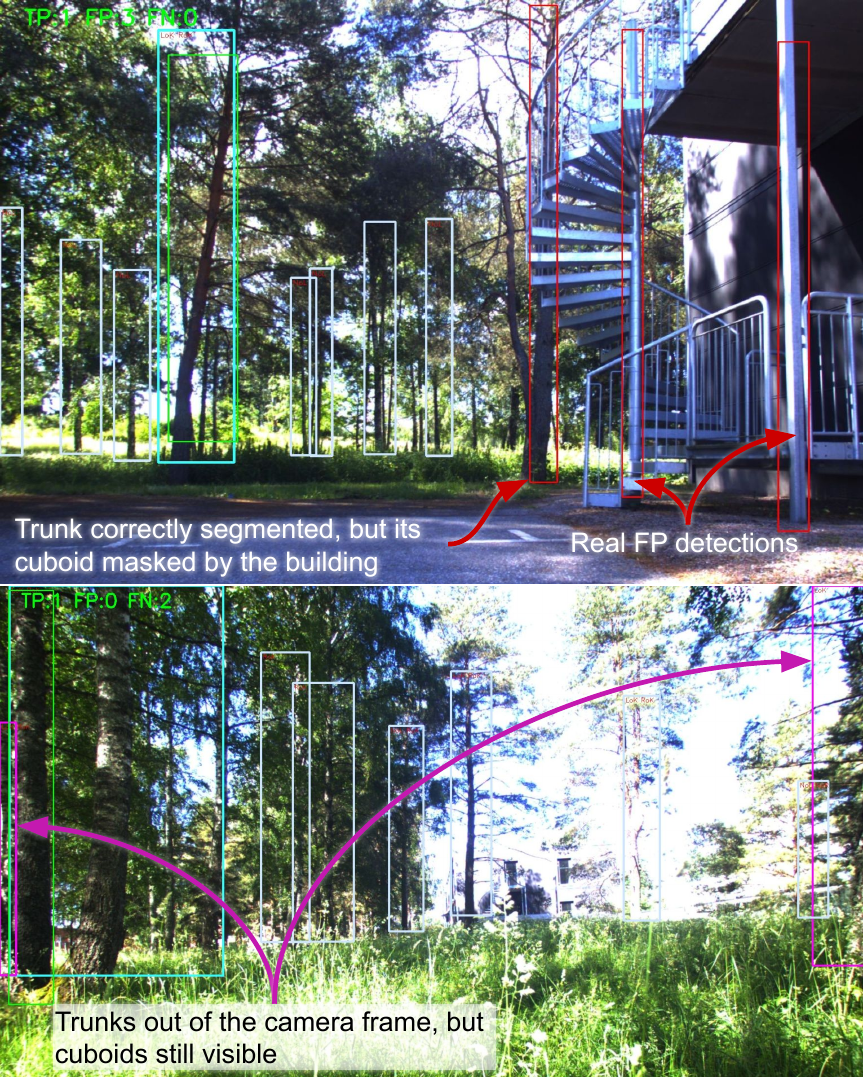}
\caption{Examples of annotation fail cases. (top) Tree trunk annotation cuboid is masked by the building cuboid. (bottom) Partially visible cuboids cause incorrect false negative detections.}
\label{fig:fails}
\end{figure}

In the current dataset, the 3D cuboids have been chosen as 3D labels for the practical impossibility of manual radar point cloud labelling.
The choice allowed us to easily transfer the labels from the lidar reference map to the radar point clouds.
However, this choice has also introduced several problems.
Fig.~\ref{fig:fails} shows cases where the simple-geometry annotation fails in real conditions.
In the top part, a tree trunk has been correctly segmented by the RGB network, its ground-truth cuboid however lies behind the building cuboid and is masked out.
This is however incorrectly considered as a false-positive case.
In the bottom image, we see edges of two cuboids (painted violet) that are still visible, even though the original tree trunks are already out of the frame.
Such a case is incorrectly considered a false negative detection.
Moreover, the trunk annotations may include nearby branch or canopy points, introducing label noise at the boundary of the trunk class.
The segmentation baseline was both trained and evaluated against these labels, so the reported metrics are self-consistent but may not reflect performance against a tighter trunk definition. 
Refining the annotation system — for instance through cylindrical fits that more closely follow the stem geometry or explicit point cloud segmentation as in \cite{malladi2025icra} — will reduce label noise and enable more precise evaluation.
This however requires finding efficient ways to create such annotations and also to distribute the labels to other modalities while respecting their specificities.
For example, radar measurements contain points that lie outside the physical objects that cause them.

Finally, the available RGB segmentation model, by design, performs best up to \SI{10}{\m}, yet the tree trunks and other objects are clearly identifiable much further away.
To provide fairer comparison between the range-sensing modalities and the camera, the annotations need to be enhanced by occlusion information.

\section{CONCLUSIONS}

We have presented a multi-modal forest dataset comprising co-registered 4D imaging radar, lidar, RGB camera, IMU, and RTK-GNSS data, recorded in two seasonal conditions and accompanied by shared 3D cuboid annotations and per-tree diameter estimates.
To our knowledge, this is the first publicly available forestry dataset that enables direct comparison of semantic segmentation performance across radar, lidar, and vision using consistent 3D labels.

Our baseline experiments with MinkowskiUNet demonstrate that high-resolution imaging radar can achieve competitive performance to lidar for dominant classes such as ground and canopy, while still lagging behind on geometrically fine structures such as tree trunks.
The cross-modality analysis conditioned on RGB trunk detection highlights complementary strengths and weaknesses of the three sensing modalities, especially under vegetation occlusion and varying distance.
Furthermore, the diameter-stratified evaluation reveals segmentation performance variation across tree sizes.
This result motivates creation of more precise annotation tools and segmentation models to better capture the variability of shapes and sizes of objects in natural environments.

Beyond segmentation, the dataset supports research in mapping, localization, and multi-sensor fusion under canopy conditions, aided by RTK-GNSS-referenced maps and consistent calibration across modalities.
By releasing the dataset, annotation tools, and baseline evaluation, we aim to facilitate reproducible evaluation and to stimulate further research into radar-based perception for field robotics.
We believe that robust multi-modal sensing, including 4D radar, will play a key role in enabling reliable autonomous operation in forests and other challenging natural environments.

\section*{ACKNOWLEDGMENT}
This work was supported by the EU Horizon Europe Framework Programme (RaCOON project, ID: 101106906) and by Knowledge Foundation (KK-stiftelsen) through the Synergy project TeamRob under grant 20210016. The 3D labels for this dataset were created using the tools from Segments.ai who generously provided a free academic license.

\bibliographystyle{IEEEtran}
\bibliography{bibtex/radar_forest_dataset}

\vfill\pagebreak

\end{document}